\documentclass[10pt,twocolumn,letterpaper]{article}

\usepackage{cvpr}              

\usepackage{multirow}
\usepackage[accsupp]{axessibility}  

\definecolor{cvprblue}{rgb}{0.21,0.49,0.74}
\usepackage[pagebackref,breaklinks,colorlinks,allcolors=cvprblue]{hyperref}

\def\paperID{*****} 
\def\confName{CVPR}
\def\confYear{2026}

\title{
PersistGS: Differentiable Physics for Object Permanence\\in 4D Gaussian Splatting
}

\author{Adrian Ramlal \quad
John S. Zelek\\
University of Waterloo\\
{\tt\small \{adrian.ramlal, jzelek\}@uwaterloo.ca}
}

\begin{document}

\twocolumn[{
\maketitle
\begin{center}
\captionsetup{type=figure}
\includegraphics[width=1\textwidth]{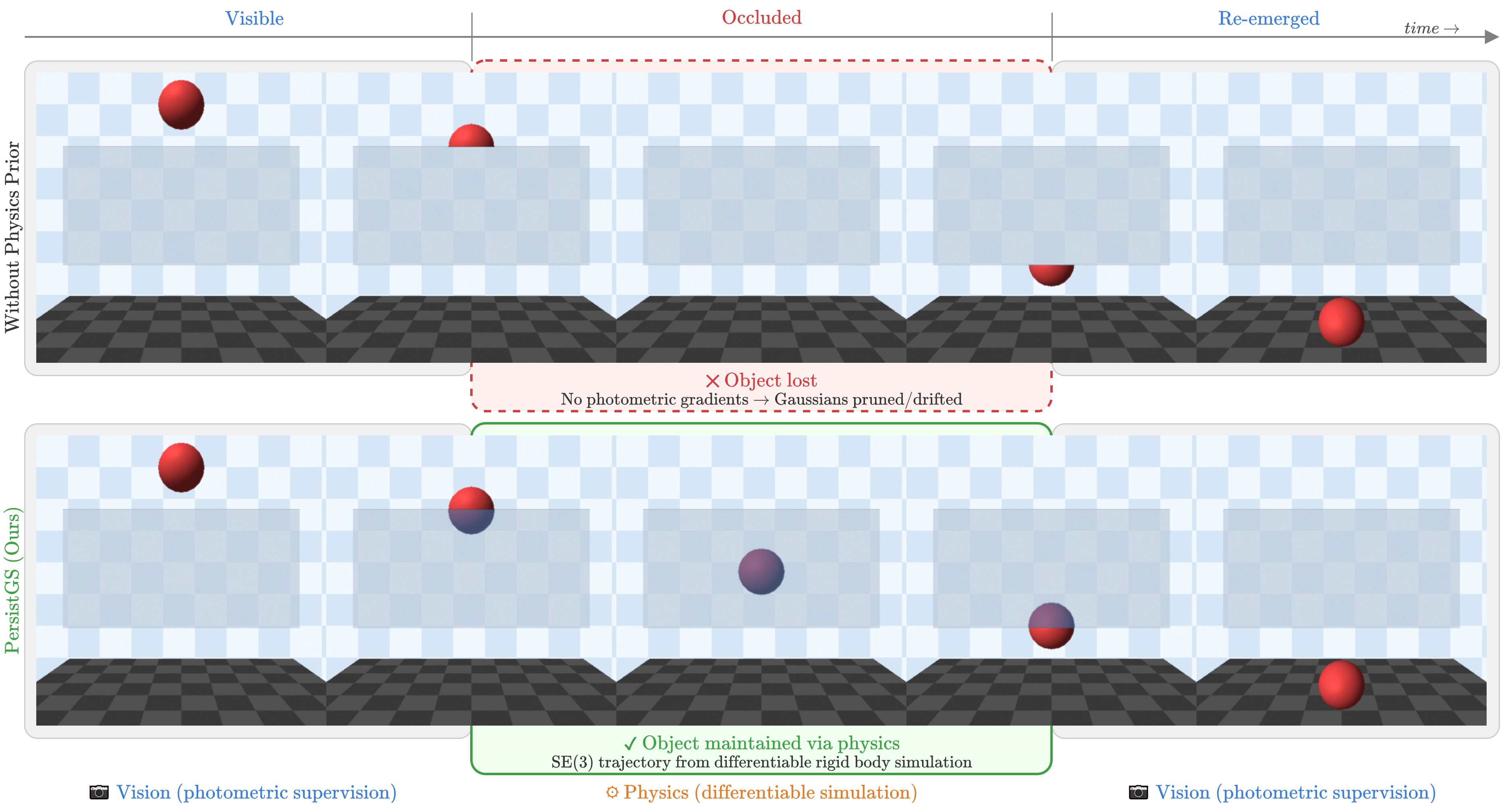}
\captionof{figure}{Object permanence through physics. A ball falls past an occluder (opaque, but rendered translucent here for visualization). Without a physics prior (top), the object's Gaussians receive no photometric gradients during occlusion and are lost. PersistGS (bottom) maintains the object throughout by positioning its Gaussians along an SE(3) trajectory predicted from pre-occlusion observations via differentiable rigid body simulation.}
\label{fig:teaser}
\end{center}
}]

\begin{abstract}




Dynamic 3D Gaussian Splatting (3DGS) methods reconstruct time-varying scenes from synchronized multi-camera video using photometric supervision. When a moving object becomes fully occluded from all training cameras, this supervision vanishes: the Gaussians representing it receive no gradient signal and degrade. Existing approaches to incomplete observations in neural reconstruction rely on learned generative priors that prioritize visual plausibility over physical correctness.

We propose \textbf{PersistGS}, a method that restores object permanence during occlusion by coupling differentiable rigid body simulation with 3D Gaussian Splatting. Our approach decomposes the scene into per-object Gaussians and collision meshes, estimates friction and velocity from the observed pre-occlusion trajectory via differentiable simulation, and uses the resulting SE(3) trajectory to position object Gaussians throughout the occlusion period. Because the predicted trajectory satisfies the governing equations of rigid body dynamics, it faithfully captures contact events (bounces, friction-based deceleration, direction changes) that kinematic extrapolation cannot model. We introduce a centroid silhouette loss that isolates positional gradients from appearance noise, yielding 40\% lower trajectory error than photometric supervision. We evaluate using cameras withheld from training that observe the object during its occlusion. Experiments on synthetic scenes show that PersistGS outperforms constant velocity extrapolation by $+2.46$~dB PSNR and comes within $0.19$~dB of a ground-truth trajectory upper bound.

\end{abstract}    

\section{Introduction}

3D Gaussian Splatting (3DGS)~\cite{kerbl2023gaussian} has become a leading representation for photorealistic scene reconstruction from multi-view images. Several extensions reconstruct dynamic scenes from synchronized multi-camera video by learning per-Gaussian deformation fields~\cite{wu20244dgs,yang2024deformable}, sparse control-point trajectories~\cite{huang2024scgs}, or per-object SE(3) poses~\cite{yan2024street,luiten2024dynamic}. These methods produce high-fidelity reconstructions when all scene elements are continuously observed across training views.

A fundamental challenge arises when observations are temporally incomplete. When a moving object passes behind a static occluder and becomes invisible from all training cameras, the Gaussians representing it receive no photometric gradients. Without gradients, these Gaussians are pruned, drift, or collapse, and the reconstruction must recreate the object from scratch upon re-emergence. The representation fails to maintain \emph{object permanence}~\cite{tokmakov2021permanence}: the principle that objects continue to exist when unobserved.

Recent work addresses incomplete observations through generative priors. Diffusion models hallucinate plausible content for unobserved regions~\cite{spotlesssplats2025}, and generative object priors complete multi-object 4D reconstructions through occlusion~\cite{chu2025genmojo,chu2024dreamscene4d}. While these approaches produce visually plausible results, they do not guarantee physical correctness. For a dynamic object moving behind an occluder, the question is not what the object might look like from an unobserved angle, but \emph{where the object is} during the unobserved interval, a problem that requires dynamical reasoning rather than appearance hallucination.

Physics provides a natural prior for this problem. Given the object's trajectory before occlusion, a rigid body simulator can predict its trajectory through the occlusion period, including contact interactions with the environment. Unlike kinematic extrapolation, physics correctly handles bounces, friction-based deceleration, and direction changes at contact. Unlike generative priors, the resulting trajectory is physically correct by construction: it satisfies Newton's laws and the contact constraints of the scene geometry.

We propose PersistGS, a method that integrates differentiable rigid body simulation with 3DGS to achieve faithful 4D reconstruction through temporal occlusion. Our approach rests on a natural decomposition: vision provides \emph{appearance} from observed frames, while physics provides \emph{position} during unobserved frames. We decompose the scene into per-object Gaussians and collision meshes, estimate physical parameters through a differentiable simulation and rendering loop, and apply the resulting SE(3) trajectory to position the object's Gaussians throughout the occlusion period. Upon re-emergence, photometric supervision seamlessly resumes.

Our contributions are:
\begin{enumerate}
    \item A method that uses differentiable rigid body simulation as a physics prior for 4D Gaussian Splatting, estimating physical parameters from observed frames and applying the resulting SE(3) trajectory to maintain object Gaussians through temporal occlusion.
    \item A centroid silhouette loss that decouples positional supervision from appearance quality, yielding 40\% lower trajectory error than photometric supervision, combined with an observability-aware curriculum for joint friction and velocity estimation.
    \item Experiments demonstrating that physics-based reconstruction through occlusion outperforms kinematic baselines by $+2.46$~dB PSNR and approaches ground-truth quality ($0.19$~dB gap), with ablations characterizing the interplay between estimation accuracy, occlusion duration, and reconstruction fidelity.
\end{enumerate}




\section{Related Work}

\subsection{Dynamic Gaussian Splatting}

3DGS~\cite{kerbl2023gaussian} represents scenes as anisotropic Gaussians optimized through differentiable rasterization. Extensions to dynamic scenes learn per-Gaussian deformations via spatiotemporal features~\cite{wu20244dgs}, continuous deformation fields~\cite{yang2024deformable}, or sparse control points with SE(3) transforms~\cite{huang2024scgs}. Dynamic 3D Gaussians~\cite{luiten2024dynamic} tracks per-Gaussian positions with rigidity constraints, and Shape of Motion~\cite{wang2024shapeofmotion} decomposes monocular video into SE(3) motion bases, the formalism our method adopts. For rigid objects, Street Gaussians~\cite{yan2024street} applies per-timestep SE(3) poses to per-object Gaussian sets, and Gaussian Grouping~\cite{ye2024gaussiangrouping} enables object-level decomposition.

Several methods explicitly separate static and dynamic scene components, a design central to our pipeline. Ex4DGS~\cite{lee2024ex4dgs} separates static and dynamic Gaussians with keyframe-based motion interpolation, SP-GS~\cite{wan2024spgs} clusters Gaussians into superpoints as group-level motion primitives, and DrivingGaussian~\cite{zhou2024drivinggaussian} models each moving object individually via a composite dynamic Gaussian graph. These decomposed architectures share our motivation of avoiding deformation field learning from limited views, but none addresses what happens when an object leaves the observable field entirely.

\subsection{Reconstruction Under Incomplete Observation}

When observations are spatially sparse, 3DGS suffers from elongation artifacts and overfitting. Mip-Splatting~\cite{yu2024mipsplatting}, DropGaussian~\cite{park2025dropgaussian}, and DNGaussian~\cite{li2024dngaussian} address this through anti-aliasing, dropout regularization, and monocular depth priors respectively. We compose these techniques to handle our clustered camera distribution.

For temporally incomplete dynamic scenes, recent methods employ learned priors. 4DGS in the Wild~\cite{kim2024wild4dgs} applies diffusion and depth priors to uncertain regions in monocular 4DGS. GenMOJO~\cite{chu2025genmojo} uses object-centric diffusion priors with occlusion-aware splatting, and DreamScene4D~\cite{chu2024dreamscene4d} lifts monocular video to 4D via amodal completion. A parallel line of work addresses the spatial variant: Amodal3R~\cite{wu2025amodal3r} reconstructs complete 3D objects from partially visible observations, and pix2gestalt~\cite{ozguroglu2024pix2gestalt} uses diffusion models for amodal segmentation. These generative approaches reason about \emph{what} the object looks like from unobserved angles; our work reasons about \emph{where} the object is during unobserved time intervals.

For transient occlusion in static scenes, RobustNeRF~\cite{sabour2023robustnerf} and SpotLessSplats~\cite{spotlesssplats2025} suppress distractors that appear inconsistently across views. Our problem is the inverse: maintaining a \emph{persistent} object that becomes \emph{temporarily unobserved}.

\subsection{Physics-Informed Neural Reconstruction}

Differentiable simulation has been coupled with neural rendering for system identification and dynamics generation. GradSim~\cite{jatavallabhula2021gradsim} introduced end-to-end parameter estimation through differentiable physics and rendering. PAC-NeRF~\cite{pacnerf2023} estimates material properties via differentiable MPM simulation, and DANO~\cite{dano2023} estimates rigid body properties from NeRF density fields with differentiable contact modeling. PhyRecon~\cite{ni2024phyrecon} demonstrates that physics priors improve neural surface reconstruction even for static scenes.

Within the Gaussian framework, most existing work couples continuum simulation with 3DGS for deformable dynamics. PhysGaussian~\cite{xie2024physgaussian} established the SE(3) Gaussian transform recipe we adopt. Subsequent methods extend this paradigm: OmniPhysGS~\cite{lin2025omniphysgs} introduces per-Gaussian learnable constitutive models, GIC~\cite{gic2024} and Spring-Gaus~\cite{zhong2024springgaus} integrate spring-mass and continuum formulations into Gaussian kernels, NeuMA~\cite{cao2024neuma} combines a physics prior with learned residual corrections for material identification, Feature Splatting~\cite{qiu2024featuresplatting} uses language queries to assign material properties for simulation, and PhysTwin~\cite{jiang2025phystwin} reconstructs and simulates deformable objects from real-world video. These methods all target \emph{deformable} materials via continuum mechanics.

Fewer methods address rigid body dynamics. SDF-Sim~\cite{rubanova2024sdfsim} learns rigid body simulation over implicit shapes from visual observations. Vid2Sim~\cite{chen2025vid2sim} reconstructs simulation-ready Gaussians with physical properties from video. Embodied Gaussians~\cite{abouchakra2024embodied} and Splatting Physical Scenes~\cite{moran2025splatting} build physics-rendering world models from robot data, and POGS~\cite{yu2025pogs} maintains persistent object Gaussians for tracking via feature matching. All of these methods use physics for forward dynamics, material estimation, or state tracking. We use physics for a distinct purpose: as a \emph{reconstruction} prior that fills temporal observation gaps with dynamically consistent trajectories.

\subsection{Object Permanence in Vision}

Object permanence, the understanding that objects continue to exist when unobserved, has been studied primarily in tracking. Tokmakov et~al.~\cite{tokmakov2021permanence} introduced it as an explicit inductive prior for multi-object tracking, using recurrent networks to predict trajectories of fully occluded objects. Vysics~\cite{vysics2025} fuses vision with contact-rich physics for shape reconstruction under occlusion, though it uses physics to resolve geometric ambiguity rather than to predict temporal trajectories.

Our work extends object permanence from tracking to reconstruction: rather than maintaining a 2D bounding box through occlusion, we maintain a complete 3D Gaussian representation positioned by a physics-predicted SE(3) trajectory. To our knowledge, PersistGS is the first method to combine differentiable rigid body simulation with Gaussian Splatting for object permanence through temporal occlusion.

\section{Method}
\label{sec:method}

\begin{figure*}[t]
    \centering
    \includegraphics[width=1\linewidth]{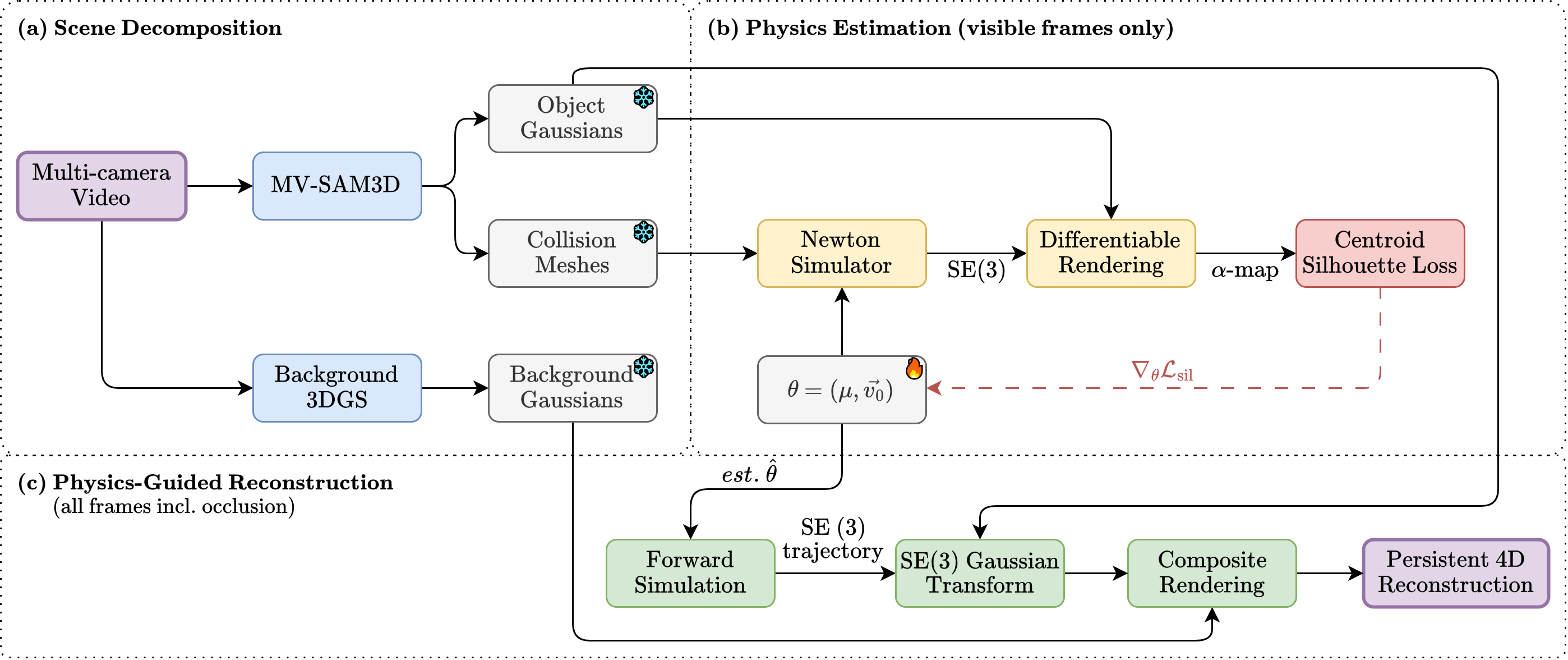}
    \caption{PersistGS pipeline. \textbf{(a)~Scene Decomposition} extracts per-object Gaussians and collision meshes via MV-SAM3D, and trains background Gaussians separately. All representations are frozen after this stage. \textbf{(b)~Physics Estimation} simulates candidate parameters $\theta = (\mu, \mathbf{v}_0)$ through Newton, renders the alpha channel of the positioned object Gaussians, and minimizes a centroid silhouette loss. Only $\theta$ is optimized. \textbf{(c)~Dynamic Reconstruction} applies the estimated SE(3) trajectory to the frozen object Gaussians and composites with the frozen background for the final 4D reconstruction.}
    \label{fig:pipeline}
\end{figure*}

Given synchronized multi-camera video of a dynamic scene, we consider the setting where a rigid object becomes fully occluded from all training cameras and later re-emerges. Our goal is a 4D reconstruction in which the object persists through occlusion with correct geometry and preserved appearance. Our pipeline has three stages: scene decomposition (\S\ref{sec:decomposition}), physics parameter estimation (\S\ref{sec:estimation}), and physics-guided reconstruction (\S\ref{sec:reconstruction}). Fig.~\ref{fig:pipeline} provides an overview.

\subsection{Scene Decomposition}
\label{sec:decomposition}

We use a decomposed representation: static background Gaussians (${\sim}$25K) and a separate per-object Gaussian model (${\sim}$511K from MV-SAM3D), composed in a single rasterization pass with automatic depth ordering. We evaluated this against a joint 4D Gaussian Splatting approach~\cite{wu20244dgs} and found the decomposed architecture outperforms by $+6.9$~dB, because 4DGS severely overfits with only 5 training cameras. The decomposed design avoids learning a deformation field from sparse views by keeping the background static and applying rigid SE(3) transforms to the object.

We reconstruct the scene at a reference frame (5~frames before the first occlusion) where all objects are visible.

\textbf{Object Gaussians.} The dynamic object is reconstructed using MV-SAM3D, which produces both a Gaussian representation and a collision mesh from posed multi-view images. The mesh provides collision geometry for the simulator.

\textbf{Background Gaussians.} The static environment is reconstructed using standard 3DGS with an inverse-mask weighted loss that excludes the object region.

\textbf{Sparse-view regularization.} Training cameras are clustered to produce consistent occlusion. We apply the Mip-Splatting 3D smoothing filter~\cite{yu2024mipsplatting}, DropGaussian dropout~\cite{park2025dropgaussian} ($p{=}0.5$), and depth supervision from Depth Anything V2~\cite{yang2024depthanythingv2} ($\lambda_{\text{depth}} {=} 0.05$).

\subsection{Physics Parameter Estimation}
\label{sec:estimation}

We estimate friction and initial velocity from the visible trajectory using a differentiable simulation and rendering loop.

\textbf{Parameterization.} We optimize friction $\mu$ (in $\log_{10}$ space) and initial velocity $\mathbf{v}_0 = (v_x, v_y, v_z)$: four free parameters. The remaining contact parameters are fixed: mass $m{=}1.0$, contact stiffness $k_e{=}10^5$, friction stiffness $k_f{=}10^3$, and damping $k_d{=}10^2$. Visual trajectory observations constrain the net forces on the object but not the individual contact parameters that produce those forces. Specifically, $k_e$ and mass enter the contact force equation only as a ratio ($a = k_e \delta / m$), making them strictly degenerate. The remaining stiffness parameters ($k_f$, $k_d$) interact with $\mu$ in governing friction buildup rate and energy dissipation during contact, making their joint estimation from trajectory data alone poorly conditioned. We therefore retain only the two parameters with the most direct and independent influence on the observable trajectory: $\mu$ (which sets the Coulomb friction limit) and $\mathbf{v}_0$ (which determines starting momentum).

\textbf{Simulator.} NVIDIA Newton~\cite{macklin2022warp,newton2025} simulates the object as a rigid body interacting with the scene's contact surfaces via a semi-implicit solver with penalty-based contact at 8 substeps per frame. The simulator outputs per-frame SE(3) poses (translation and quaternion). Warp's tape-based reverse-mode automatic differentiation provides gradients through the full trajectory, including contact events.

Enabling effective gradient flow through contact dynamics required two modifications to Newton's contact kernels. First, the default friction model uses a hard $\min(\cdot)$ clamp that creates zero-gradient regions when the friction force saturates at the Coulomb limit. We replace this with a smooth harmonic combination $f = (k_f v_t \cdot f_c) / (k_f v_t + f_c + \epsilon)$, where $f_c$ is the Coulomb limit, ensuring non-zero gradients in all regimes. Second, Newton's default material averaging ($k_e = 0.5(k_{e,\text{body}} + k_{e,\text{shape}})$) routes only 50\% of the gradient to the optimizable shape material parameters. We bypass this averaging, yielding approximately $50{\times}$ improvement in gradient magnitude for contact parameters.

\textbf{Centroid silhouette loss.} Pixel-wise photometric losses combine two sources of error: the object's position (governed by physics) and its appearance (fixed from the pre-occlusion reconstruction). Because the object Gaussians are frozen, their spherical harmonic coefficients may not produce correct color from novel viewpoints, injecting appearance noise into physics gradients.

We instead supervise with a centroid silhouette loss that isolates position from appearance. The object Gaussians are transformed to the simulated SE(3) pose and rendered to produce an alpha map. The rendered centroid is compared against the ground-truth mask centroid:
\begin{equation}
    \mathcal{L}_{\text{sil}} = \| \mathbf{c}_{\text{render}} - \mathbf{c}_{\text{gt}} \|_2^2, \quad \mathbf{c} = \frac{\sum_{i} \alpha_i \mathbf{p}_i}{\sum_{i} \alpha_i}
    \label{eq:centroid}
\end{equation}
This loss provides a global basin of attraction: even when predicted and ground-truth silhouettes do not overlap, the centroid displacement yields a non-zero gradient toward the correct position. It also produces friction gradients approximately $100{\times}$ larger than photometric loss (Sec.~\ref{sec:ablations}), because centroid displacement is a first-order function of position, while pixel-wise RGB differences arise only at silhouette boundaries. Because centroid displacement is dominated by translation for any object whose radius is small relative to the scene, the loss provides effective positional gradients for physics parameters regardless of object geometry.

\textbf{Observability-aware curriculum.} Pre-contact frames constrain velocity but carry no friction information ($\partial\mathcal{L}/\partial\mu {=} 0$ in free flight), while post-contact frames constrain both. We exploit this structure:
\begin{enumerate}
    \item \emph{Phase~1} ($60$ iter.): velocity from pre-contact frames, warm-started from finite differences, friction frozen.
    \item \emph{Phase~2} ($60$ iter.): friction from a post-contact frame window, velocity frozen.
    \item \emph{Phase~3} ($80$ iter.): joint refinement at $0.3{\times}$ learning rate.
\end{enumerate}
We run 5 random initializations per scene and select the lowest-loss result.

\subsection{Physics-Guided Reconstruction}
\label{sec:reconstruction}

The estimated parameters drive a full forward simulation, producing an SE(3) trajectory through the occlusion period. Background Gaussians remain static while object Gaussians are transformed per-frame following the SE(3) recipe for Gaussian dynamics~\cite{xie2024physgaussian}: positions $\boldsymbol{\mu}' {=} \mathbf{R}_t\boldsymbol{\mu}_{\text{can}} {+} \mathbf{t}_t$, quaternions $\mathbf{q}' {=} \mathbf{q}_{R_t} {\otimes} \mathbf{q}_{\text{can}}$, and scales unchanged under rigid motion. For spherical harmonics, we apply the inverse object rotation to the viewing direction before evaluation ($\text{color} = \text{SH}(\mathbf{R}_t^{-1} \mathbf{d}_{\text{view}};\, \mathbf{C}_{\text{can}})$), which is algebraically equivalent to rotating all SH coefficients but avoids computing Wigner D-matrices.

During visible frames, per-frame residual translations are optimized over 3 passes through the visible frame set (learning rate $10^{-4}$), minimizing photometric error against training views to correct small positional errors from imperfect parameter estimation. During occluded frames, the object follows the physics trajectory purely, with residuals disabled and no gradients propagating to the object representation. At re-emergence, residual optimization and photometric supervision resume, ensuring a smooth transition back to vision-guided reconstruction. All Gaussians are composited in a single rasterization pass.

\section{Experiments}
\label{sec:experiments}

\subsection{Setup}

\textbf{Scenes.} We evaluate on three synthetic scenes simulated with NVIDIA Newton~\cite{newton2025}. Each features a rigid ball (radius 1.25) in a static environment with a ground plane and occluding wall, designed so that contact events occur during occlusion:

\begin{itemize}
    \item \textbf{ball\_fall:} Free fall with lateral drift; ground bounce behind occluder. Occlusion: 248 frames (69\% of 360). GT: $\mu {=} 0.3$, $\mathbf{v}_0 {=} (3, 0, 0)$~m/s.
    \item \textbf{ball\_bounce:} Parabolic arc; second bounce hidden behind occluder. Occlusion: 80 frames (33\% of 240). GT: $\mu {=} 0.15$, $\mathbf{v}_0 {=} (5, 0, 7)$~m/s.
    \item \textbf{ball\_roll:} Ground roll with friction deceleration behind occluder. Occlusion: 45 frames (13\% of 360). GT: $\mu {=} 0.4$, $\mathbf{v}_0 {=} (10, 0, 0)$~m/s.
\end{itemize}

Each scene uses 5 training cameras and 2 evaluation cameras (overhead and side, withheld from training) at $512{\times}512$, 60~fps.

\textbf{Baselines.} (1)~\textbf{GT trajectory} (upper bound); (2)~\textbf{Ours}; (3)~\textbf{Linear interpolation} between last-visible and first-reappearance positions (non-causal: requires the re-emergence location); (4)~\textbf{Constant velocity} with gravity but no contact model (the only causal kinematic baseline); (5)~\textbf{No physics} (object absent during occlusion).

\textbf{Metrics.} PSNR, LPIPS, and SSIM on an object-region crop from evaluation cameras during occlusion, plus 3D trajectory RMSE.

\textbf{Implementation.} All experiments use a single NVIDIA RTX~5080 GPU (16~GB). Physics estimation takes ${\sim}5.5$~min/scene (200 iterations across 3 curriculum phases, 5 random seeds). Background Gaussians are trained for 5K iterations (${\sim}$25K Gaussians). Object Gaussians are initialized from MV-SAM3D and fine-tuned for 7K iterations (${\sim}$511K Gaussians, DropGaussian rate 0.5, degree-3 spherical harmonics).

\subsection{Parameter Estimation}

Table~\ref{tab:params} reports estimated physical parameters. Friction is recovered within 10\% and velocity within 0.5~m/s on all components. Trajectory RMSE during occlusion ranges from 0.288 (ball\_roll) to 1.147 (ball\_fall).

\begin{table}[t]
\centering
\caption{Estimated vs.\ ground-truth physical parameters.}
\label{tab:params}
\setlength{\tabcolsep}{4pt}
\begin{tabular}{l l rrr}
\toprule
Scene & Param & GT & Est. & Error \\
\midrule
\multirow{4}{*}{ball\_fall}
 & $\mu$ & 0.300 & 0.290 & 3.3\% \\
 & $v_x$ & 3.000 & 2.999 & 0.001 \\
 & $v_y$ & 0.000 & $-$0.007 & 0.007 \\
 & $v_z$ & 0.000 & $-$0.045 & 0.045 \\
\midrule
\multirow{4}{*}{ball\_bounce}
 & $\mu$ & 0.150 & 0.151 & 0.7\% \\
 & $v_x$ & 5.000 & 4.993 & 0.007 \\
 & $v_y$ & 0.000 & $-$0.094 & 0.094 \\
 & $v_z$ & 7.000 & 6.994 & 0.006 \\
\midrule
\multirow{4}{*}{ball\_roll}
 & $\mu$ & 0.400 & 0.363 & 9.3\% \\
 & $v_x$ & 10.000 & 9.527 & 0.473 \\
 & $v_y$ & 0.000 & 0.033 & 0.033 \\
 & $v_z$ & 0.000 & $-$0.218 & 0.218 \\
\bottomrule
\end{tabular}
\end{table}

\subsection{Reconstruction Quality During Occlusion}

Table~\ref{tab:main} presents the primary evaluation. PersistGS achieves 17.15~dB mean PSNR on evaluation cameras during occlusion, outperforming constant velocity by $+2.46$~dB and linear interpolation by $+1.41$~dB, while approaching the ground-truth upper bound (17.34~dB, gap of $0.19$~dB). LPIPS confirms consistent improvements: 0.314 vs.\ 0.381 (linear) and 0.491 (constant velocity).

\begin{table}[t]
\centering
\caption{Reconstruction during occlusion (object-region crop, evaluation cameras withheld from training). GT trajectory is the upper bound.}
\label{tab:main}
\resizebox{\columnwidth}{!}{%
\begin{tabular}{l cccc cccc}
\toprule
& \multicolumn{4}{c}{PSNR $\uparrow$ (dB)} & \multicolumn{4}{c}{LPIPS $\downarrow$} \\
\cmidrule(lr){2-5} \cmidrule(lr){6-9}
Method & Fall & Bnc & Roll & Mean & Fall & Bnc & Roll & Mean \\
\midrule
GT (upper bnd) & 16.08 & 18.33 & 17.60 & 17.34 & .403 & .203 & .247 & .284 \\
\textbf{Ours} & \textbf{15.76} & \textbf{18.02} & \textbf{17.67} & \textbf{17.15} & \textbf{.405} & \textbf{.279} & \textbf{.257} & \textbf{.314} \\
Linear interp. & 11.98 & 17.75 & 17.50 & 15.74 & .500 & .339 & .305 & .381 \\
Const.\ velocity & 14.13 & 14.93 & 15.01 & 14.69 & .555 & .469 & .450 & .491 \\
No physics & 10.63 & 12.63 & 12.77 & 12.01 & .761 & .716 & .669 & .716 \\
\bottomrule
\end{tabular}}
\end{table}

\begin{figure*}[t]
    \centering
    \includegraphics[width=\linewidth]{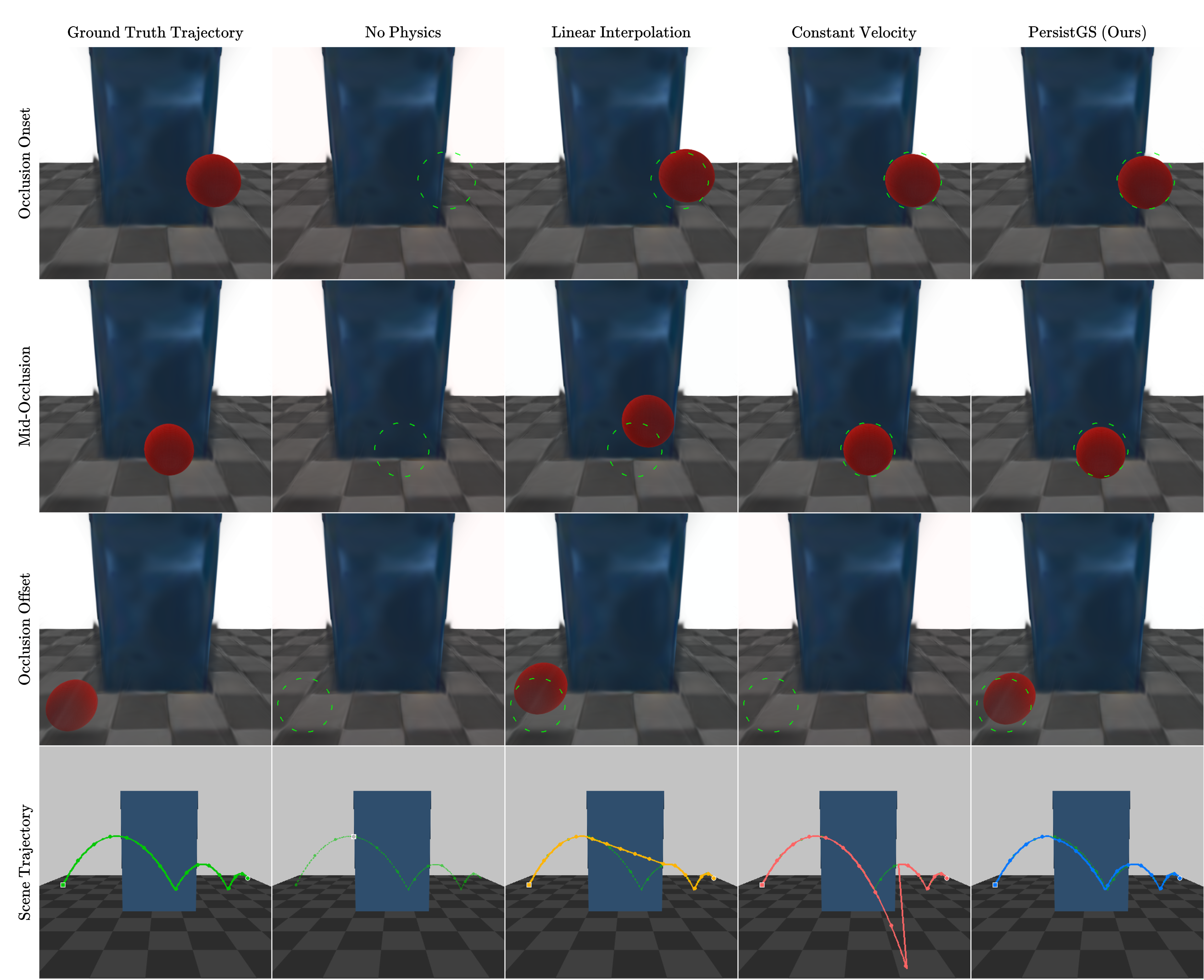}
    \caption{Qualitative results on ball\_bounce. Top three rows: renders from an evaluation camera (which sees past the occluder) at three stages of the occlusion event. Green dotted outlines indicate the ground-truth ball position. Without physics, the ball is absent; constant velocity misses the second bounce; linear interpolation follows a straight path through the nonlinear contact trajectory; PersistGS correctly tracks the physics-predicted arc. Bottom row: full-scene trajectories from a training viewpoint, with the ground-truth path (green) overlaid on each method's Gaussian reconstruction.}
    \label{fig:results}
\end{figure*}

The physics advantage is scene-dependent. On \textbf{ball\_fall} (248-frame occlusion with a ground bounce), physics outperforms interpolation by $+3.78$~dB because the nonlinear contact trajectory cannot be recovered by a straight-line path. On \textbf{ball\_bounce}, the hidden second bounce yields a $+3.09$~dB advantage over constant velocity. On \textbf{ball\_roll} (45-frame, nearly linear occlusion), interpolation performs comparably ($+0.17$~dB gap), but constant velocity overshoots by $-2.66$~dB because it ignores friction. Fig.~\ref{fig:results} shows qualitative results on ball\_bounce.

Table~\ref{tab:rmse} reports trajectory RMSE during occlusion. Constant velocity diverges on ball\_fall (8.936) and ball\_bounce (5.716), reflecting its inability to model contact interactions.

\begin{table}[t]
\centering
\caption{Trajectory RMSE during occlusion (lower is better).}
\label{tab:rmse}
\begin{tabular}{l cccc}
\toprule
Method & Fall & Bounce & Roll & Mean \\
\midrule
\textbf{Ours} & \textbf{1.147} & \textbf{0.366} & 0.288 & \textbf{0.600} \\
Linear interp. & 3.603 & 0.720 & \textbf{0.222} & 1.515 \\
Const.\ velocity & 8.936 & 5.716 & 1.427 & 5.360 \\
\bottomrule
\end{tabular}
\end{table}

\subsection{Ablations}
\label{sec:ablations}

\textbf{Centroid vs.\ photometric loss.} Table~\ref{tab:loss_ablation} compares the centroid silhouette loss (Eq.~\ref{eq:centroid}) against masked pixel-wise MSE for physics parameter estimation. Centroid achieves 40\% lower mean RMSE (0.586 vs.\ 0.984) and $+0.41$~dB higher PSNR. The advantage is strongest on ball\_fall (3.4$\times$ lower RMSE), where centroid loss produces ${\sim}100{\times}$ larger friction gradients ($|\nabla_\mu| > 10$ vs.\ ${\sim}0.05$) because centroid displacement is a first-order function of position while pixel-wise RGB differences arise only at silhouette boundaries. On ball\_bounce, where friction is unobservable from pre-contact visible frames ($\partial\mathcal{L}/\partial\mu {=} 0$ regardless of loss), both perform comparably.

\begin{table}[t]
\centering
\caption{Centroid silhouette vs.\ photometric loss for physics estimation.}
\label{tab:loss_ablation}
\resizebox{\columnwidth}{!}{%
\begin{tabular}{l cc cc cc cc}
\toprule
& \multicolumn{2}{c}{Fall} & \multicolumn{2}{c}{Bounce} & \multicolumn{2}{c}{Roll} & \multicolumn{2}{c}{Mean}\\
\cmidrule(lr){2-3}\cmidrule(lr){4-5}\cmidrule(lr){6-7}\cmidrule(lr){8-9}
Loss & RMSE$\downarrow$ & PSNR$\uparrow$ & RMSE$\downarrow$ & PSNR$\uparrow$ & RMSE$\downarrow$ & PSNR$\uparrow$ & RMSE$\downarrow$ & PSNR$\uparrow$ \\
\midrule
Centroid (ours) & \textbf{0.61} & \textbf{16.44} & 1.05 & 17.02 & \textbf{0.10} & \textbf{17.57} & \textbf{0.59} & \textbf{17.01} \\
Photometric & 2.08 & 15.59 & \textbf{0.77} & \textbf{17.24} & 0.10 & 16.96 & 0.98 & 16.60 \\
\bottomrule
\end{tabular}}
\end{table}

\textbf{Noise tolerance.} Adding i.i.d.\ noise ($\sigma$) to the ground-truth trajectory yields graceful degradation (Table~\ref{tab:noise}): approximately 1~dB per $\sigma{=}0.25$ increment. Even at $\sigma{=}2.0$, PSNR exceeds the no-physics baseline (12.01~dB). Our estimated trajectory RMSE (0.29--1.15) falls within $\sigma {\approx} 0.3$--$1.2$, confirming that the estimation accuracy is sufficient for high-quality reconstruction.

\begin{table}[t]
\centering
\caption{PSNR (dB) vs.\ trajectory noise $\sigma$ (evaluation cameras, occlusion).}
\label{tab:noise}
\begin{tabular}{c ccc}
\toprule
$\sigma$ & Fall & Bounce & Roll \\
\midrule
0.00 & 16.08 & 18.33 & 17.60 \\
0.10 & 15.07 & 17.19 & 17.09 \\
0.25 & 13.97 & 15.66 & 16.17 \\
0.50 & 12.74 & 14.30 & 15.02 \\
1.00 & 11.47 & 13.24 & 13.75 \\
2.00 & 10.92 & 12.76 & 13.06 \\
\bottomrule
\end{tabular}
\end{table}

\textbf{Occlusion duration.} The physics advantage over constant velocity grows with occlusion length: on ball\_roll, the gap widens from $+1.83$~dB (40 frames) to $+2.49$~dB (69 frames), because kinematic extrapolation accumulates error while physics respects contact constraints.

\textbf{Visible frame count.} Ball\_fall PSNR improves from 13.07~dB ($N{=}5$ visible frames) to 16.20~dB (all frames), while ball\_bounce is stable at ${\sim}17.3$~dB regardless of $N$. Complex trajectories benefit from more observations; simpler dynamics can be estimated from few frames.

\textbf{Sparse-view regularization.} Without regularization, ball\_bounce and ball\_roll degrade by $+3.0$ and $+4.2$~dB respectively on evaluation cameras, confirming that regularization is essential when training views are clustered.

\section{Discussion and Future Work}

\textbf{When does physics help?} The advantage of physics over kinematic baselines scales with both occlusion duration and contact complexity. On ball\_fall (248 frames, ground bounce during occlusion), physics outperforms even the non-causal linear interpolation by $+3.78$~dB, while on ball\_roll (45 frames, nearly linear segment), kinematic models are competitive. The noise tolerance ablation reveals a smooth, monotonic relationship between trajectory accuracy and reconstruction quality (${\sim}1$~dB per $\sigma{=}0.25$), indicating that even approximate physics estimates are preferable to kinematic assumptions when contacts are present. This suggests a practical decision criterion: physics priors are most valuable when the occluded interval contains contact events that produce nonlinear trajectory changes, and least necessary when the trajectory is approximately ballistic.

\textbf{Parameter observability.} Parameter identifiability depends on which physical interactions the cameras can observe. On ball\_bounce, where first ground contact occurs during occlusion, friction is unobservable from pre-contact visible frames regardless of loss function. The curriculum addresses this by matching each parameter to the frames where it is identifiable. Preliminary two-pass experiments, where post-occlusion observations refine the physics parameters, show up to $+0.80$~dB improvement on ball\_bounce by enabling friction estimation from re-emergence frames, suggesting that iterative refinement is a promising direction. More broadly, the identifiability analysis reveals a fundamental tension: the scenes where physics helps most (complex contacts during occlusion) are precisely those where the relevant parameters are hardest to estimate from visible frames alone.

\textbf{Causality and practical applicability.} Linear interpolation, despite competitive performance on two scenes, requires the re-emergence position, which is unavailable in any causal or predictive setting. Against constant velocity, the only causal baseline, physics provides a consistent $+2.46$~dB advantage, making PersistGS the strongest causal method evaluated. This causal property is essential for downstream applications such as robotic planning and digital twin maintenance, where an agent must reason about occluded object locations in real time without access to future observations. The decomposed architecture (static background plus per-object SE(3)) further supports such applications by providing an explicit object-level representation that can interface directly with a planner or controller.

\textbf{Architecture.} The decomposed architecture proved essential: a joint 4DGS baseline~\cite{wu20244dgs} overfits by $+6.9$~dB with only 5 training cameras, confirming that the sparse-view setting demands separation of static and dynamic components. This finding aligns with concurrent work on decomposed dynamic Gaussian representations~\cite{lee2024ex4dgs,zhou2024drivinggaussian} and suggests that physics-based trajectory supervision is most naturally integrated within object-centric architectures where each rigid body has an explicit SE(3) trajectory to constrain.

\textbf{Limitations and future work.} Spherical objects provide a clean validation of the translational component of our centroid silhouette loss, since sphere silhouettes are rotation-invariant. For asymmetric objects, the silhouette covariance (second central moment of the alpha map) captures projected orientation and offers a natural extension to jointly constrain translation and rotation without relying on appearance. The full SE(3) pipeline, including rotation-aware SH evaluation, is implemented and exercised during reconstruction. Validating on objects where orientation produces distinct visual changes is a natural next step. Extending to multi-object scenes with inter-object contact, and integrating automatic decomposition methods~\cite{ye2024gaussiangrouping} to remove the requirement for known object segmentation, are further directions.

\section{Conclusion}

We presented PersistGS, a method that uses differentiable rigid body simulation as a physics prior for 4D Gaussian Splatting, maintaining object permanence through temporal occlusion. By estimating friction and velocity from visible frames via a centroid silhouette loss and observability-aware curriculum, and positioning object Gaussians along the resulting SE(3) trajectory, PersistGS produces faithful reconstructions through observation gaps. On three scenes with contact events during occlusion, it outperforms constant velocity extrapolation by $+2.46$~dB PSNR and comes within $0.19$~dB of the ground-truth upper bound. The centroid loss yields 40\% lower trajectory error than photometric supervision, and modifications to Newton's contact kernels enable effective gradient flow through rigid body dynamics. These results establish analytical physics simulation as a principled alternative to generative priors for 4D reconstruction from temporally incomplete observations.

{
    \small
    \bibliographystyle{ieeenat_fullname}
    \bibliography{main}
}


\end{document}